\documentclass[letterpaper, 10 pt, conference]{ieeeconf}  % Comment this line out if you need a4paper

\IEEEoverridecommandlockouts                              % This command is only needed if 
\usepackage[T1]{fontenc}
\usepackage{aecompl}

\usepackage{algorithmic}
\usepackage{algorithm}

\usepackage{graphicx}
\usepackage{float}
\usepackage{booktabs}
\usepackage{bm}
\usepackage{array}
\usepackage{amsmath}
\usepackage{amssymb}
\usepackage{threeparttable}
\usepackage{url}
\usepackage{subfigure}
\usepackage{multirow}

\title{\LARGE \bf
Learning Individual Intrinsic Reward in Multi-Agent Reinforcement Learning via Incorporating Generalized Human Expertise
}

\author{Xuefei Wu$^{*1}$, Xiao Yin$^{*1}$, Yuanyang Zhu$^{\dag2}$ and Chunlin Chen$^{1}$% <-this % stops a space
\thanks{*~Equal contribution, \dag~Corresponding author.}% <-this % stops a space
\thanks{$^{1}$Xuefei Wu, Xiao Yin, and Chunlin Chen are with the Department of Control Science and Intelligent Engineering, School of Management and Engineering, Nanjing University, Nanjing 210093, China. (email: xuefeiwu,~xiaoyin@smail.nju.edu.cn, clchen@nju.edu.cn).}%
\thanks{$^{2}$Yuanyang Zhu is with the Laboratory of Data Intelligence and Interdisciplinary Innovation, School of Information Management, Nanjing University, Nanjing 210023, China. (email: yuanyangzhu@nju.edu.cn)}%
\thanks{This work was supported in part by the China Postdoctoral Science Foundation under Grant Number 2025T180877, the National Key Research and Development Program of China under Grant 2023YFD2001003, Major Science and Technology Project of Jiangsu Province under Grant BG2024041 and the Fundamental Research Funds for the Central Universities under Grant 011814380048.}
}

\begin{document}

\maketitle
\thispagestyle{empty}
\pagestyle{empty}

%%%%%%%%%%%%%%%%%%%%%%%%%%%%%%%%%%%%%%%%%%%%%%%%%%%%%%%%%%%%%%%%%%%%%%%%%%%%%%%%
\begin{abstract}
Efficient exploration in multi-agent reinforcement learning (MARL) is a challenging problem when receiving only a team reward, especially in environments with sparse rewards.
A powerful method to mitigate this issue involves crafting dense individual rewards to guide the agents toward efficient exploration.
However, individual rewards generally rely on manually engineered shaping-reward functions that lack high-order intelligence, thus it behaves ineffectively than humans regarding learning and generalization in complex problems. 
To tackle these issues, we combine the above two paradigms and propose a novel framework, LIGHT (Learning Individual Intrinsic reward via Incorporating Generalized Human experTise), which can integrate human knowledge into MARL algorithms in an end-to-end manner.
LIGHT guides each agent to avoid unnecessary exploration by considering both individual action distribution and human expertise preference distribution.
Then, LIGHT designs individual intrinsic rewards for each agent based on actionable representational transformation relevant to Q-learning so that the agents align their action preferences with the human expertise while maximizing the joint action value.
Experimental results demonstrate the superiority of our method over representative baselines regarding performance and better knowledge reusability across different sparse-reward tasks on challenging scenarios.
\end{abstract}

%%%%%%%%%%%%%%%%%%%%%%%%%%%%%%%%%%%%%%%%%%%%%%%%%%%%%%%%%%%%%%%%%%%%%%%%%%%%%%%%
\section{Introduction}
Cooperative multi-agent reinforcement learning~(MARL) is an important branch in the field of artificial intelligence~(AI), playing a crucial role in sequential challenging decision-making problems, such as in autonomous driving~\cite{car}, sensor networks~\cite{zhang2011coordinated,ghorbani2019interpretation} and robotics control~\cite{huttenrauch2017guided}.
Centralized training with decentralized execution~(CTDE) paradigm has gained substantial attention in cooperative MARL that aims to facilitate agent cooperation by providing global state information during training and executing only based on local observations during execution~\cite{qpd, rode, roma}. 
Recent research has witnessed extensive investigation into value decomposition methods under the paradigm of CTDE.
Since the advances in these MARL approaches~\cite{vdn, qmix, qplex, mixrts,liu23be}, well-designed auxiliary rewards are indispensable to guide agent collaboration or competition.
Unfortunately, many cooperative multi-agent tasks currently only offer common team rewards~\cite{busoniu2008comprehensive}.

In real-world multi-agent systems, sparse team rewards present a significant challenge~\cite{sadeghlou2014dynamic}.
Existing algorithms rely on dense-reward environments for guiding efficient cooperation strategies~\cite{wong2021multiagent}, but these conditions rarely hold in real-world scenarios. 
In MARL, handling sparse-reward environments often involves enhancing agent exploration, a common approach demonstrated to be effective in various tasks~\cite{liu2021cooperative, mahajan2019maven}. 
However, relying solely on exploration can be insufficient for determining which specific actions trigger rare non-zero rewards.
It highlights the need for methods that promote discovery and aid in identifying and reinforcing these critical action-reward connections.
Individual intrinsic rewards can be a promising solution to this problem~\cite{liir, maser,wang2022individual}, typically by distributing a combination of team and individual rewards among agents~\cite{xu2023haven, xu2020hierarchical}. 
It can change agents' learning objectives, potentially leading to unexpected behaviors divergent from desired team outcomes.
An important line of work that leverages human knowledge is a promising method to improve the learning process efficiently~\cite{fischer2019dl2,9403986,9970401,zhang2020kogun}.
However, a major challenge in leveraging human knowledge is how to obtain the representation of the provided knowledge.

To solve this problem, we propose a novel method called LIGHT, Learning Individual Intrinsic reward via Incorporating Generalized Human experTise, which can plug into the value decomposition algorithms to promote the learning efficiency, especially in sparse-reward setting tasks under the widely-used assumption of CTDE.
Our key insight is that instead of using human knowledge to directly guide the agent to interact with the environment, we integrate human knowledge to produce the intrinsic reward to induce the agent to achieve better exploration.
Specifically, at each time step, LIGHT learns a parameterized intrinsic reward function by considering the action distribution and human preference that outputs an intrinsic reward for each agent to implicitly induce diversiﬁed behaviors.

We evaluate LIGHT on two representative benchmarks: Level-Based Foraging~(LBF) and StarCraft Multi-Agent Challenge~(SMAC), where empirical results demonstrate our method's superior performance over other baselines.
We conduct further component studies to show the effectiveness of individual intrinsic reward incorporated with generalized human expertise for agent learning, which confirms that each component is a key part of LIGHT.
We also find that the behavior of LIGHT obtains better alignment with human knowledge, indicating it provides an efficient method to incorporate human preference into the learning process.

\section{Preliminaries}
\textbf{Multi-agent Markov Decision Process.}
A fully cooperative multi-agent task can be modeled as an extension of a Decentralized Partially Observable Markov Decision Process (Dec-POMDP), which is defined by a tuple $G = \langle \mathcal{N}, \mathcal{S}, \mathcal{A}, \mathcal{P}, \Omega, O, r, \gamma \rangle$, where $\mathcal{N}$ represents a finite set of agents with $\mathcal{N} \equiv \{1, 2, \ldots, n\}$, $s \in \mathcal{S}$ denotes the global state of the environment.
At each time step $t$, each agent $i \in \mathcal{N}$ selects an action $a_i \in \mathcal{A}$ to formulate a joint action $\boldsymbol{a} \equiv [a_i]_{i=1}^{n} \in \mathcal{A}^n$.
The joint action leads to a shared reward according to the reward function $r(s, \boldsymbol{a})$ and a transition to a new state based on the transition probability function $s' \sim \mathcal{P}(\cdot | s, \boldsymbol{a})$.
Given the partial observability, each agent $i$ receives an individual observation $o_i \in \Omega$, associated with the observation probability function $O(o_i | s, a_i)$.
The action-observation history for each agent $i$ is denoted as $\tau_i \in \mathcal{T} \equiv (\Omega \times \mathcal{A})^*$, and the joint action-observation history is $\boldsymbol{\tau} \in \mathcal{T}^n$.
The objective for all agents is to find an optimal joint policy  $\boldsymbol{\pi} = \langle \pi_1, \ldots, \pi_n \rangle$ to maximize the expected cumulative reward $\mathbb{E}[\sum_{t=0}^{\infty} \gamma^{t}r^{t}]$, where $\gamma \in [0,1)$ is a discount factor.

\textbf{Centralized Training with Decentralized Execution.} CTDE is a prevalent paradigm in the MARL, where each agent learns a policy only on its own action observations, and the centralized critic provides a global perspective, offering gradient updates that are informed by the joint state and action space.
A promising enhancement of the CTDE framework is the application of value decomposition. This technique enables agents to individually learn utility functions that collectively optimize the joint action-value function, thereby providing a clear framework for credit assignment among agents.
For the integrity of multi-agent value decomposition methods, adherence to the Individual-Global-Max (IGM) principle is essential.
To ensure consistency~\cite{qtran} for multi-agent value decomposition methods, it should satisfy the IGM principle:
\begin{equation}
	\label{dsdsadad}
	\operatorname{argmax}_{\boldsymbol{a}} Q_{tot}(\boldsymbol{\tau}, \boldsymbol{a})
	=\left(\begin{array}{c}
		\operatorname{argmax}_{a_{1}} Q_{1}\left(\tau_{1}, a_{1}\right) \\
		\vdots \\
		\operatorname{argmax}_{a_{n}} Q_{n}\left(\tau_{n}, a_{n}\right)
	\end{array}\right),
\end{equation}

\textbf{Individual Intrinsic Reward.}
Individual intrinsic rewards have become pivotal in MARL settings, particularly where extrinsic rewards are sparse.
The LIIR~\cite{liir} incorporates individual intrinsic rewards into the Actor-Critic~(AC) algorithm by designing an attentive reward mechanism. 
It can be formally defined as
\begin{equation}
	R_{i, t}^{\mathrm{proxy}}=\sum_{l=0}^{\infty} \gamma^l\left(r_{t+l}^{\mathrm{ex}}+\lambda r_{i, t+l}^{\mathrm{in}}\right),
\end{equation}
where $\lambda$ is a tunable hyperparameter that balances the influence of global extrinsic rewards and the intrinsic reward.
The introduction of $\lambda$ allows for a dynamic adjustment between learning from the environment and fostering behaviors motivated by the agent’s own experiences.
The resultant proxy value function for each agent is expressed as
\begin{equation}
	V_i^{\text {proxy }}\left(s_{i, t}\right)=\mathbb{E}_{a_{i, t}, s_{i, t+1}, \cdots}\left[R_{i, t}^{\text {proxy }}\right],
\end{equation}
where $a_i$ is the action space of each agent $i$ at time step $t$.
Then the proxy value function is applied to optimize the agents' policy. 
ICQL~\cite{ICQL} introduces a local uncertainty measure to enhance learning in decentralized agents through intrinsic motivation. 
This technique fosters a nuanced understanding of the environment by encouraging exploration through uncertainty-driven intrinsic rewards, thereby complementing the LIIR framework’s strategy for balancing intrinsic and extrinsic rewards.

\begin{figure*}[htb]
	\centering
	\includegraphics[width=2\columnwidth]{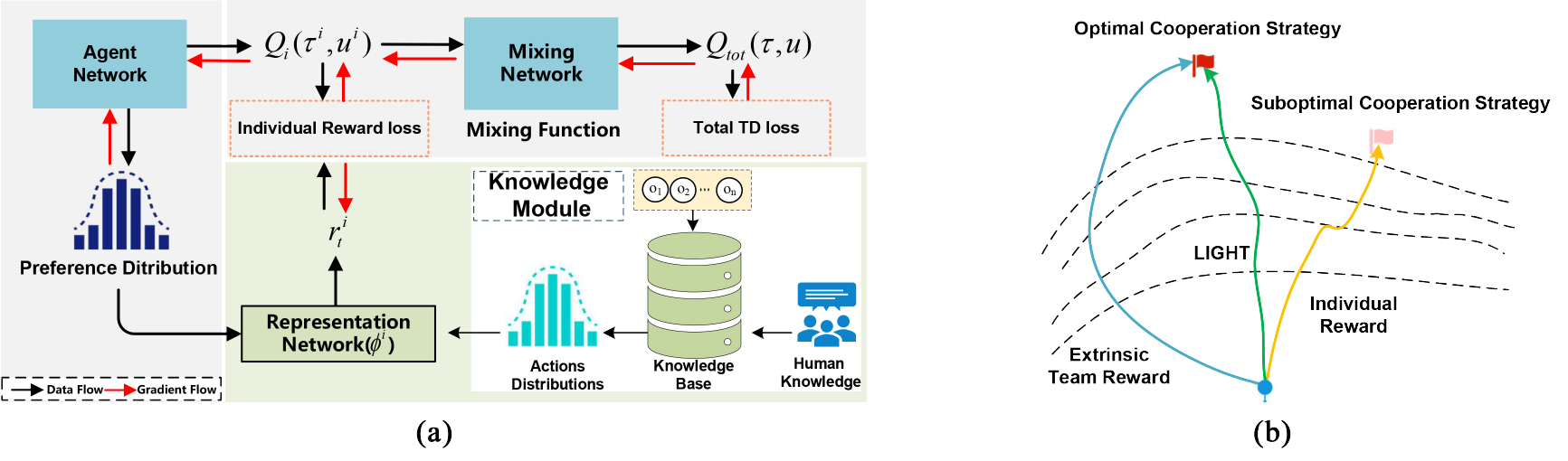}
	\vskip -0.15in
	\caption{The framework of LIGHT. 
		(a) In the knowledge module, action distributions are generated through a knowledge base considering states and human knowledge, and then it is compared with a preference distribution generated through the agent network by a representation network. The resulting intrinsic reward is denoted as $r_t^i$. 
		Then the agent network will output $Q_i\left(\tau^i, u^i\right)$, which are mixed into a $Q_{tot}(\tau, u)$ by a mixing network.
		Additionally, the parameters are optimized using both the total TD loss and the individual reward loss, allowing the framework to be trained end-to-end.
		(b) On the sparse-reward scenarios, the extrinsic reward is hard to guide the agents to get the optimal strategy, and then the individual intrinsic reward is added to the global reward together with the extrinsic reward.
		By maximizing the global reward, the agents can improve the learning efficiency and reduce the number of exploration steps.
	}
\vskip -0.26in
	\label{framework}
    
\end{figure*}

\section{Method}
In this section, we propose a novel end-to-end cooperative MARL framework called LIGHT, solving the MARL with sparse reward effectively via leveraging human knowledge to generate individual intrinsic rewards.
This section introduces the methodology of LIGHT. 
We begin with our motivation and then provide a detailed explanation of the implementation of LIGHT.

%\vspace{-0.1cm}
\subsection{Motivation}
Sparse-reward scenarios are typical in RL applications, where agents may not have enough information to develop an optimal behavior and may learn to exploit suboptimal but easily accessible solutions.
In CTDE, each agent acts independently with local observability and receives only the factorization global reward.
This shared reward structure complicates the learning of cooperative policies, as it can be tough to discern which actions contribute to the success of the group.
It makes it difficult for an algorithm to successfully learn a cooperative team policy in such a setting.
One could also consider a manual specification of dense rewards.
However, designing a useful reward function is notoriously difficult and time-consuming.
This naturally leads to the fundamental question: Can we design informative rewards that will guide the agent to efficiently explore and accelerate the agent's learning process?

Recalling the learning process of humans, they rarely approach the acquisition of new skills in a vacuum.
They adeptly draw upon a wealth of prior knowledge derived from analogous tasks to formulate an initial strategy.
Human cognition is characterized by the ability to extract explainable, task-solving heuristics that exhibit a degree of generalizability across related domains.
If this essence of human knowledge could be distilled into the fabric of RL agents by encoding logical inferences into the neural architectures that underpin their learning processes.
These agents could bypass the initial trial-and-error phase and embark immediately on the refinement of effective strategies.

One natural solution is to integrate human knowledge to produce the reward.
Following this idea, we introduce our LIGHT framework as illustrated in Fig.~\ref{framework}, which synthesizes human-derived insights into intrinsic rewards, thereby enhancing exploration and optimizing collective outcomes.
Within this framework, intrinsic rewards are dynamically generated by integrating a parameterized, knowledge-informed reward model, bolstering overall learning efficiency.
During training, agents receive their local observations and produce their individual action-value distribution.
Then, this estimate is contrasted with a corresponding distribution shaped by human expertise, yielding a knowledge-based intrinsic reward signal.
This reward signal is tailored to each agent's local observations and embedded understanding, enabling LIGHT to circumvent the inefficiencies of uninformed exploration typical of early training stages.
Consequently, LIGHT provides stable and coherent intrinsic rewards that effectively guide each agent, facilitating a more robust and efficient credit assignment process in MARL.

\subsection{The Practical Implementation of LIGHT}
Based on the previous analysis, we will give the implementation details of LIGHT.
The key idea of LIGHT is incorporating human knowledge to identify superior and inferior behaviors during the learning process and encourage agents to achieve superior behaviors while avoiding sticking to inferior ones.
This implicit guidance of learning by human knowledge is imposed in a progressive manner as the learning proceeds.
First, we select a few logical rules that align with human knowledge, where logic rules are extracted from the offline MARL data via decision trees.
Then, we define the individual intrinsic reward for each agent $i$ at time step $t$ as the negative Euclidean distance between the current individual action distributions and the preference action distribution of human knowledge.
This individual intrinsic reward will be used to update the mixing network.
By maximizing the global reward, the agents can implicitly share human knowledge to improve learning efficiency through parameter sharing among agents.

\textbf{Human knowledge.}
Here, to obtain human knowledge, we extract logic rules from offline data~\cite{offline} with decision tree techniques.
To leverage human knowledge, a major challenge is obtaining a representation of the provided knowledge.
Under most circumstances, the provided knowledge is imprecise and uncertain, and even covers only a small part of the state space.
The hard rules, born from the bivalent framework, starkly contrast with the adaptable nature of human knowledge.
Such rules are ill-suited to model the rich, often indistinct patterns of thought that humans employ, underscoring the need for a more malleable and encompassing approach to knowledge representation.
It needs one to mirror the intricate tapestry of human understanding faithfully.
To relieve the uncertainty and imprecision, we transform the selected logic rules into soft logic rules, where the probability of the decision process of soft logic rules is attached to probability $p\in[0,1]$ rather than the prior probability $p=1$.

\begin{figure}[ht]
	\centering
	\includegraphics[width=0.9\columnwidth]{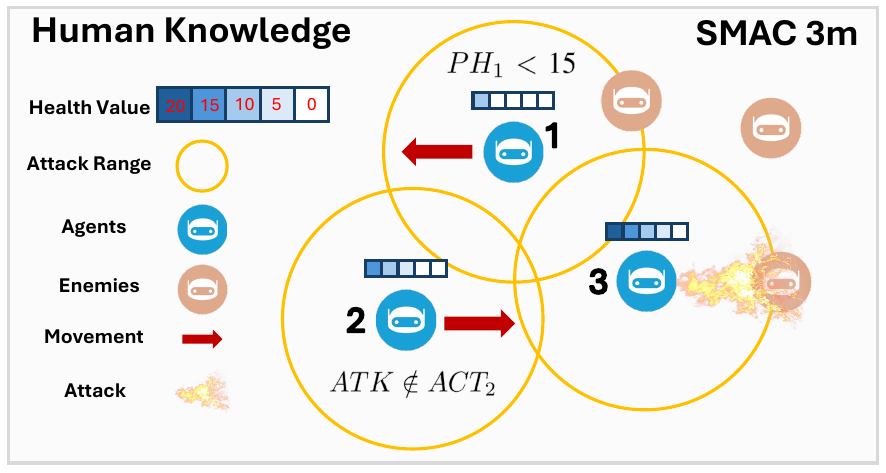}
    \vskip -0.1in
	\caption{Example of a task with the rule about human knowledge.}
	\label{rule}
    \vskip -0.3in
\end{figure}

To clearly understand the rule, we provide an example of the rule regarding human knowledge on 3m map in SMAC and summarize the rule accordingly. At each time step $t$, we obtain each agent’s health point $PH$ and action set $ACT=\{M, ATK, N, ST\}$, where M represents movement in four directions, $ATK$ is the $attack$ action, $N$ is the $none$ action, and $ST$ represents the death of agents. As shown in Fig.~\ref{rule}, when the health point of agent-1 falls below 15 and when actionable options of agent-2 exclude attack, the probability distribution over available actions shifts significantly toward movement behaviors.
\begin{algorithm}[t]
	\caption{\textbf{Rule for human knowledge on 3m map.}}	
	\textbf{Input:} {All agents' health points $PH$ and action set $ACT = \{M, ATK, N, ST\}$, where $M=\{North, South, East, West\}$}. \\
	\textbf{Output:}{The probability $p$.}
	
	\begin{algorithmic}[1]
		\FOR {each agent $i$}
		\IF {$PH_i$ $<$ 15 or $ATK \notin ACT_i$}
		\STATE {Get the probability $p$, where $\operatorname{argmax}_{a_i} p = M$}
		\ENDIF
		\ENDFOR
	\end{algorithmic}
\end{algorithm}
%\vspace{-0.2cm}

\textbf{Intrinsic Reward.} 
The joint policy learns from trajectories sampled by individual policies and delineates the pursuit of maximized group utility.
The individual policy $\pi_i$, in a dynamic state of recalibration, is based on their sampling distributions in response to the iterative refinements of the team policy.
The purpose of using the intrinsic reward is to optimize and maximize the global reward, as shown in Fig.~\ref{framework}-(b).
The sparse team rewards make it hard to guide the joint policy to the optimal policy, like the blue line.
Through the factorization of team utility, each agent rarely receives beneficial signals to update its policy towards the optimal policy (orange line).
To this end, we introduce the intrinsic reward, which incorporates human knowledge to implicitly guide individual agents to achieve better exploration.

At each time step $t$, the agent $i$ receives its intrinsic reward by computing the negative Euclidean distance between its action distribution and soft logic rules.
When receiving its local observation $o_i^t$ at time step $t$, we compute the intrinsic reward $r_t^i$ as
\begin{equation}
	\label{eq4}
	r_t^{i}=-\left\|\phi^i\left(\mathcal{H}({o_t^i})\right)-\phi^i\left(A_t^i\right)\right\|_2, t=1,...,T,
\end{equation} 
where $\phi^i(\cdot)$ denotes an operable representational transformation, $\mathcal{H}({o_t^i})$ is the action distribution of the soft logic rules, and $A_t^i$ is a value distribution of each agent.
Each agent tries to reach its returns while maximizing the team's returns.
It can also be regarded as the difference in actions between human preferences and individual agents.

We redefine the reward function to quantify the contribution of individual agents to the success of the team, combining the environment's extrinsic reward with added intrinsic incentives:
\begin{equation}
	R_t=r_t^{e x}+\lambda \frac{1}{N} \sum\nolimits_{i=1}^N r_t^{i, j},
\end{equation}
where $R_t$ is used to update the mixed network parameters $\theta$.
To relieve the sparsity of extrinsic rewards, intrinsic rewards are added to make the mixing parameters non-trivially updated at each time step.

\textbf{Overall Learning.} 
Under the training execution framework of CTDE, LIGHT learns by sampling a multitude of transitions from a replay buffer, and the loss function for the mixing network parameter is represented as
\begin{equation}
	L(\theta)=\left(R_t+\gamma \max _{u^{\prime}} Q_{\theta^{-}}^{\text {tot }}\left(s_{t+1}, u^{\prime}\right)-Q_\theta^{\text {tot }}\left(s_t, u_t\right)\right)^2,
\end{equation}
where $\theta_i^{-}$ is the parameter of the target network for the mixing network.
The individual Q-values update the intrinsic reward value of each agent. 
The overall learning objective is to minimize the following loss:
\begin{equation}
	L_i\left(\theta_i\right)=\left[r_t^i+\gamma \max_{u^i} Q_{\theta_i^{-}}^i\left(o_{t+1}^i, u^i\right)-Q_{\theta_i}^i\left(o_t^i, u_t^i\right)\right]^2,
\end{equation}
where $L_i$ represents the individual loss value of each agent, and $r_t^i$ stands for the intrinsic reward.
The total loss function used in this work is expressed as follows:
\begin{equation}
	L=L_{T D}(\theta)+\lambda_K L_i\left(\theta_i\right),
\end{equation}
where $\lambda_K$ is denoted as a coefficient set for individual loss.

\section{Experiments}
In this section, we evaluate LIGHT on the widely-used and challenging tasks over Level-Based Foraging~(LBF)~\cite{lbf} and StarCraft Multi-Agent Challenge~(SMAC)~\cite{smac} benchmarks, where SMAC includes dense-reward and sparse-reward settings.
We compared with five representative MARL algorithms: MASER~\cite{maser}, LIIR~\cite{liir}, VDN~\cite{vdn}, QMIX~\cite{qmix} and QTRAN~\cite{qtran}.
Then, we conduct ablation studies on LIGHT to better understand each component’s effect.
To ensure fair evaluation, we conducted all experiments with five random seeds, and the results are plotted using mean $\pm$ std.

\section{Experimental Environment Settings}

\subsection{Level-Based Foraging~(LBF)}
LBF is a sparse reward and mixed cooperative-competitive environment, where agents collect randomly-scattered food items by navigating a grid world.
Each agent navigates a 10 $\times$ 10 grid world while observing a 5 $\times$ 5 sub-grid centered at its current position.
The food collection is successful when the sum of the near-by agents' levels exceeds the food's level.
Then, agents can receive rewards that are equal to the level of the food they collect, divided by their level.
We construct two scenarios with different quantities of agents and food to evaluate the performance of all methods, including 4 agents with 2 food (4-agent $\&$ 2-food) and 3 agents with 3 food (3-agent $\&$ 3-food).

\subsection{StarCraft Multi-Agent Challenge~(SMAC)}
SMAC simulates intricate scenarios from the acclaimed real-time strategy game StarCraft, providing a robust environment for validating our proposed methodologies.
Each agent in our setup is equipped with a local observation vector drawing from a wealth of tactical data points, including the proximity, placement, vitality, shield status, and classification of both allied and adversarial units.
A noteworthy feature of the simulation is the dynamic shield regeneration that activates after a designated duration of non-combat status, alongside an armor mechanic that necessitates depletion before any reduction in an agent's health pool can occur.
Our experiments are conducted against a formidable opposition in-game built-in AI, which operates at a difficulty level of 7.

\textbf{The Reward Setting for Two Scenarios.} 
In our investigation, we benchmark our LIGHT framework against leading state-of-the-art MARL algorithms across both dense-reward and sparse-reward environments.
The specific configurations for these reward structures are detailed in Table~\ref{paramater2}.
The dense-reward scenario adheres to the conventional reward paradigm, aligning with the standard reward mechanism implemented in the PyMARL framework.
In this setup, agents receive frequent feedback, with rewards dispensed for both significant milestones and incremental progressions during the course of an episode.
Conversely, the sparse-reward scenario in SMAC experiments aligns with the parameters set forth in the MASER framework~\cite{maser}.
Within this challenging environment, the reward signal is markedly reduced, with agents receiving feedback exclusively upon the culmination of the task or in response to critical in-game events such as the elimination of enemy units or the loss of allied ones.
This comparative study aims to scrutinize the efficacy of our LIGHT framework in learning and decision-making processes under contrasting reward densities, thereby offering comprehensive insights into its performance relative to established MARL benchmarks in strategic game settings.
\vspace{-0.2cm}

\begin{table}[t]
	\vspace{10pt}
	\caption{The configurations of reward settings.}
	\centering
	\small
	\begin{tabular}{ccc}
		%\centering
		\toprule
		& Dense reward & Sparse reward \\
		\hline 
		Win & +200 & +200 \\
		Enemy's death & +10 & +10 \\
		Ally's death & -5 & -5 \\
		Enemy's health & -Enemy's remaining health & - \\
		Ally's health & +Ally's remaining health & - \\
		Other elements & +/- with other elements & - \\
		\bottomrule
	\end{tabular}
	\label{paramater2}	
\end{table}

\begin{table*}[t]
	\centering
	\caption{The configurations of hyperparameter settings for LIGHT and other baselines.}
	\small
	\begin{tabular}{ccccccc}
		\toprule
		& LIGHT & MASER & LIIR & QMIX & VDN  & QTRAN \\
		\hline 
		Buffer Size & 5000 & 5000 & 32 & 5000 & 5000 & 5000 \\
		Batch Size & 32 & 32 & 32 & 32 & 32 & 32 \\
		Test Interval~(SMAC) & 2000 & 2000 & 2000 & 2000 & 2000 & 2000 \\
		Test Interval~(LBF) & 1000 & 1000 & 1000 & 1000 & 1000 & 1000 \\
		Test Episodes & 32 & 32 & 32 & 32 & 32 & 32 \\
		Optimizer & RMSProp & RMSProp & RMSProp & RMSProp & RMSProp & RMSProp \\
		Agent Runner & episode & episode & parallel & episode & episode & episode \\
		Learning Rate & 0.0005 & 0.0005 & 0.0005 & 0.0005 & 0.0005 & 0.0005 \\
		TD Discounted Factor & 0.99 & 0.99 & 0.99 & 0.99 & 0.99 & 0.99 \\
		Start Exploration Rate & 1 & 1 & 0.5 & 1 & 1 & 1 \\
		End Exploration Rate & 0.05 & 0.05 & 0.01 & 0.05 & 0.05 & 0.05 \\
		Epsilon Anneal Step & 50000 & 50000 & 50000 & 50000 & 50000 & 50000 \\
		Target Update Interval & 200 & 200 & 200 & 200 & 200 & 200 \\
		Mixing Embed Dimension & 32 & 32 & - & 32 & - & 64 \\
		\bottomrule 
	\end{tabular}
	\label{paramater1}
\end{table*}

\section{Hyperparameter Settings}
The hyperparameters of configurations follow the source code provided by the authors while keeping it consistent across all baselines for fairness.
The detailed hyperparameters for LIGHT and other baselines on LBF and SMAC can be found in Table~\ref{paramater1}.
In addition, LIGHT employs greedy action selection, and $\lambda_K$ is set to $0.02$.
%\vspace{-0.35cm}

\begin{figure}[t]
	\centering
	\includegraphics[width=1\columnwidth]{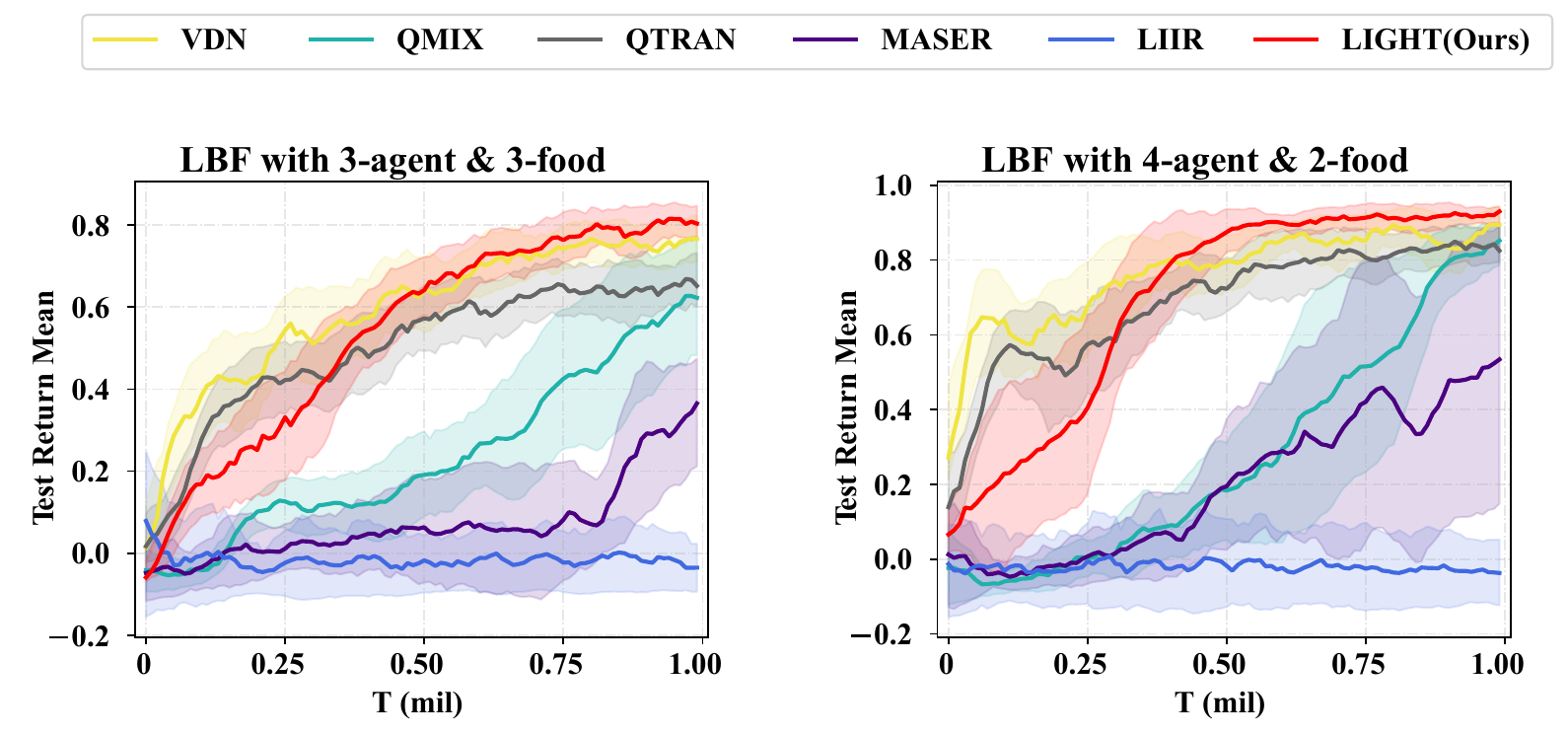}
    \vskip -0.15in
	\caption{Performance comparison with different baselines on two constructed scenarios of LBF.}
	\label{performance_lbf}
    \vskip -0.2in
\end{figure}

\begin{figure}[t]
	\centering
	\subfigure[Intrinsic reward curves]{
		\includegraphics[width=0.16\textwidth]{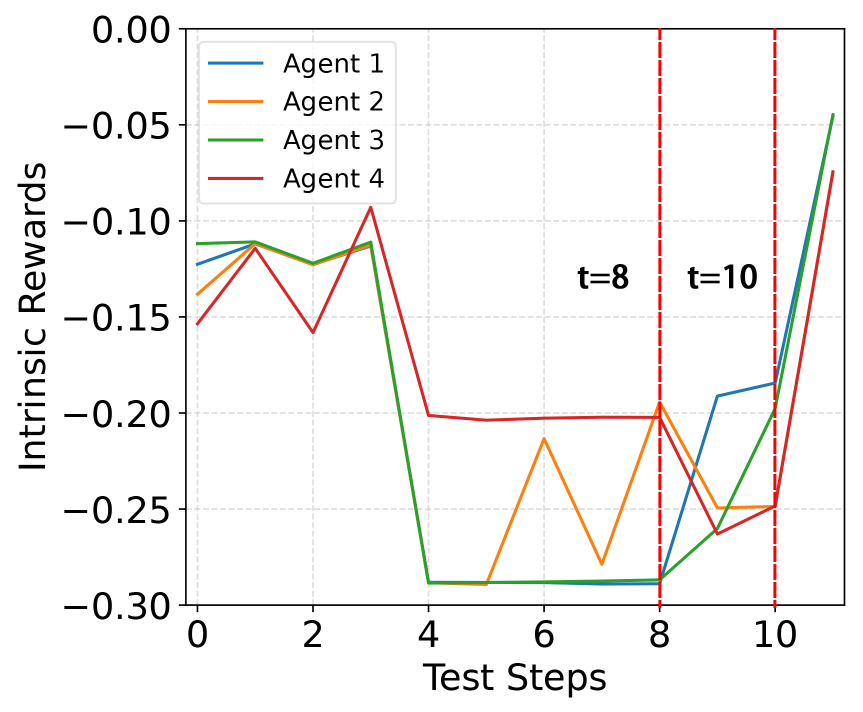}
		\label{lbf-re}
	}
	\subfigure[t=8]{
		\includegraphics[width=0.13\textwidth]{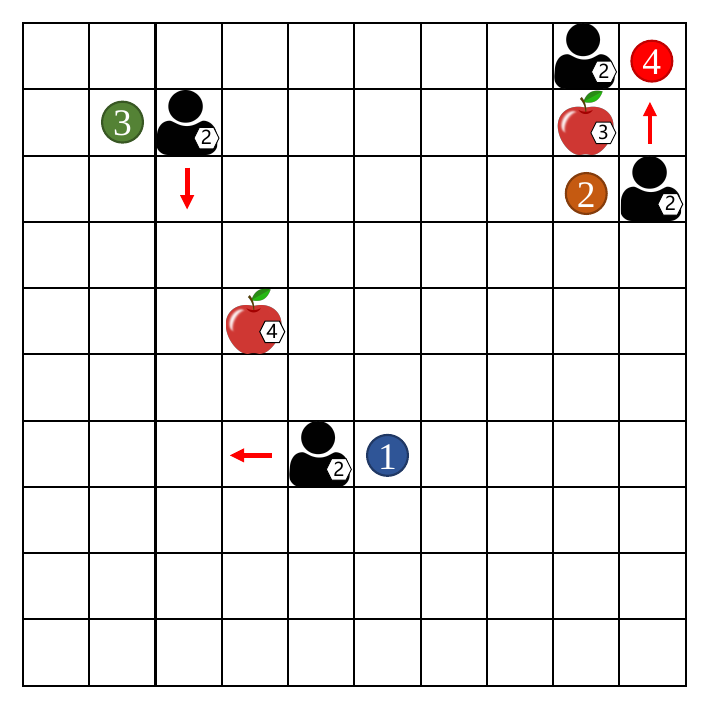}
		\label{lbf-t8}
	}
	\subfigure[t=10]{
		\includegraphics[width=0.13\textwidth]{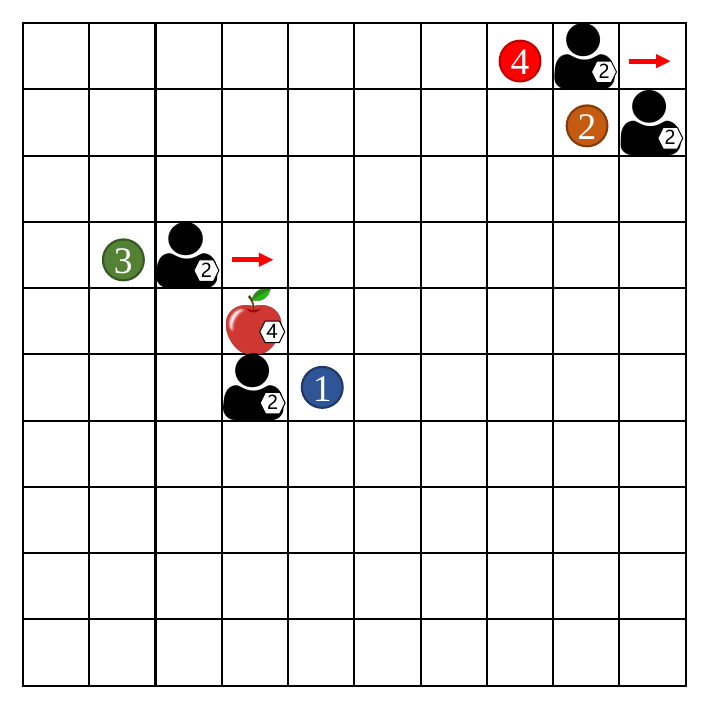}
		\label{lbf-t10}
	}
    \vskip -0.05in
	\caption{An example of the intrinsic reward curves and some keyframes on LBF with 4-agent $\&$ 2-food.}
	\label{LBF}
    \vskip -0.3in
\end{figure}

\begin{figure*}[ht]
	\centering
	\includegraphics[width=2\columnwidth]{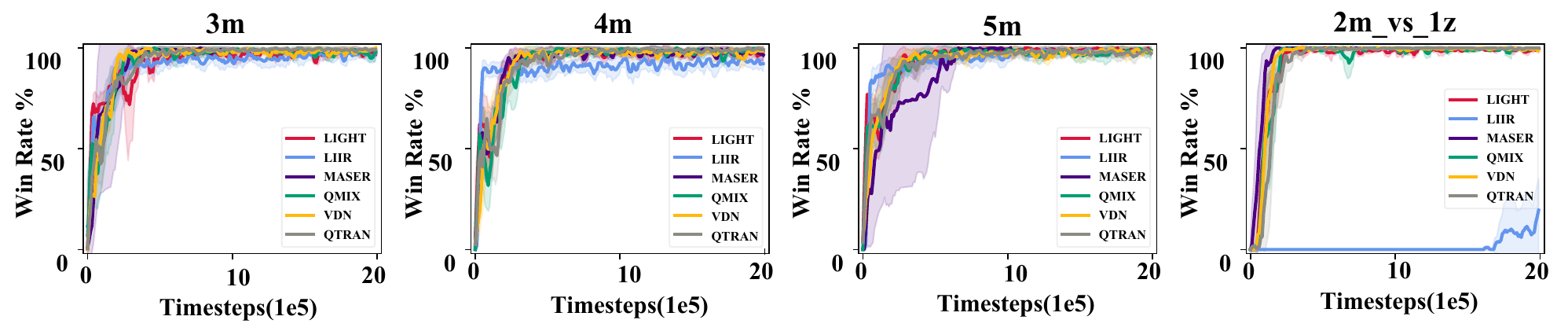}
    \vskip -0.15in
	\caption{Performance comparison with baselines on dense-reward setting scenarios.}
	\label{normal}
    \vskip -0.15in
\end{figure*}

\begin{figure*}[ht]
	\centering
	\includegraphics[width=2\columnwidth]{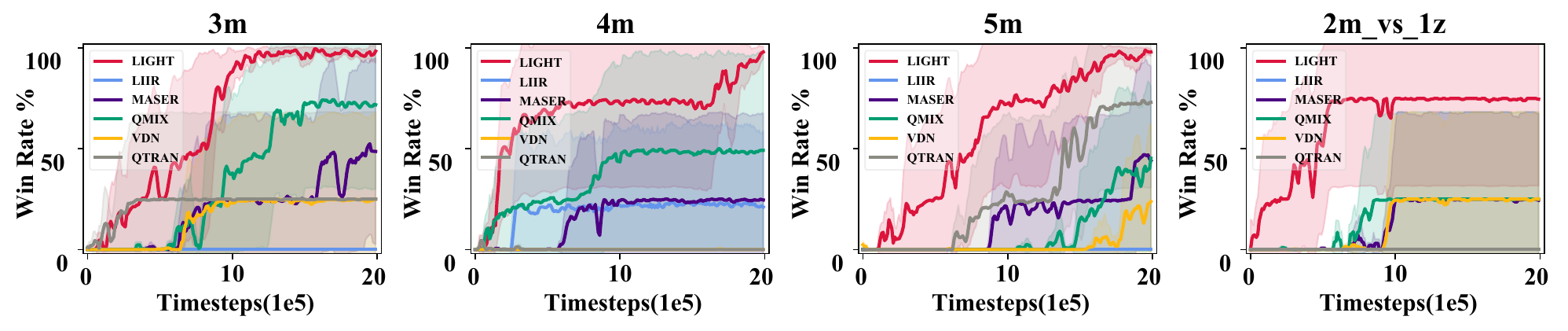}
    \vskip -0.15in
	\caption{Performance comparison with baselines on sparse-reward setting scenarios.}
	\label{sparse}
    \vskip -0.2in
\end{figure*}

\begin{figure}[ht]
	\centering
	\subfigure[{\fontsize{7}{1}\selectfont Intrinsic reward curves}]{
		\includegraphics[width=0.152\textwidth]{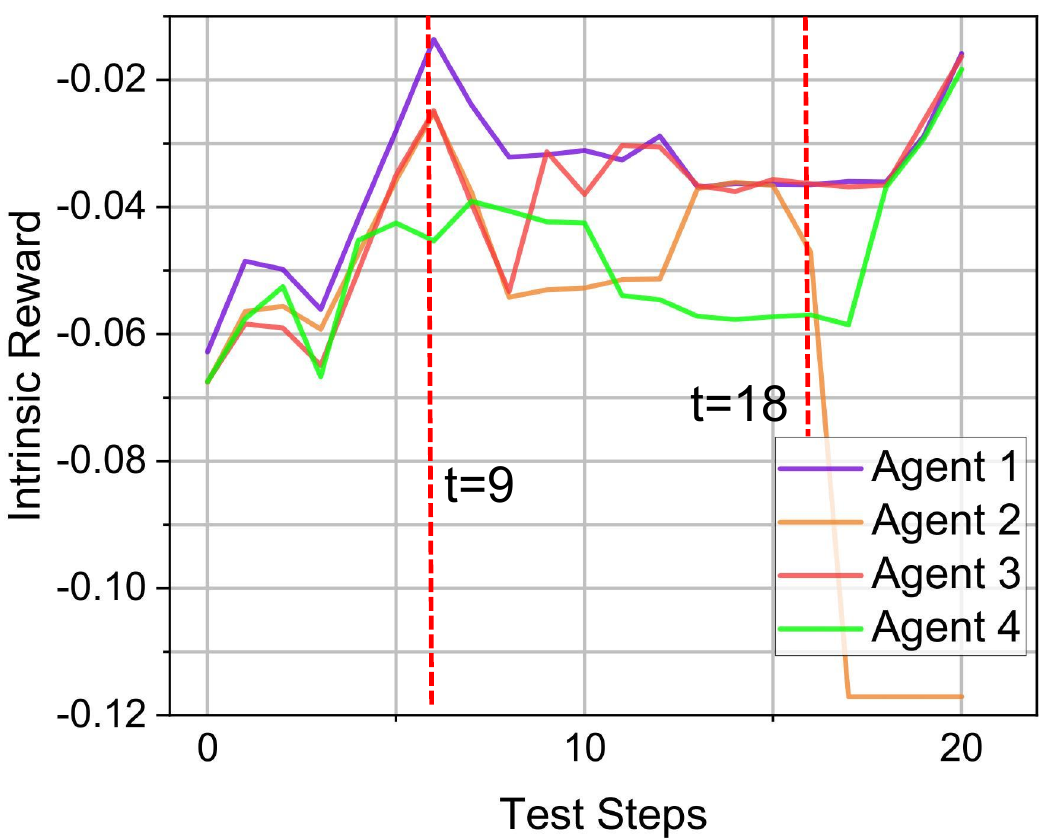}
	}
	\subfigure[t=9]{
		\includegraphics[width=0.14\textwidth]{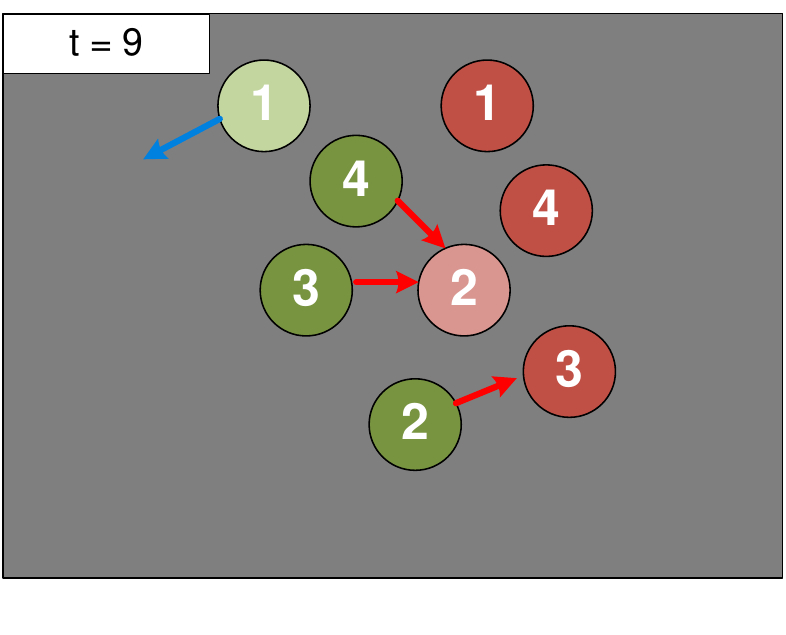}
	}
	\subfigure[t=18]{
		\includegraphics[width=0.14\textwidth]{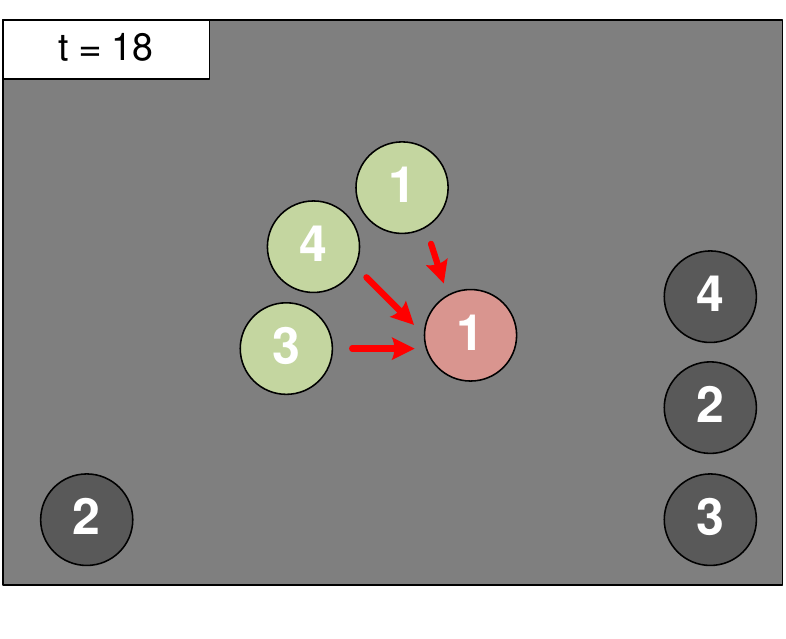}
	}
    \vskip -0.05in
	\caption{An example of the intrinsic reward curves and auxiliary illustration on 4m map.
		Green circles and red circles represent allies and enemies, respectively, where the darker color indicates the higher health value of the agent.
		Gray circles indicate that the agent was killed.
		Blue arrows represent $move$, red arrows represent $attack$.}
	\label{4m_intrinsic_reward}
    \vspace{-0.6cm}
\end{figure}

\begin{figure*}[ht]
	\centering
	\includegraphics[width=2\columnwidth]{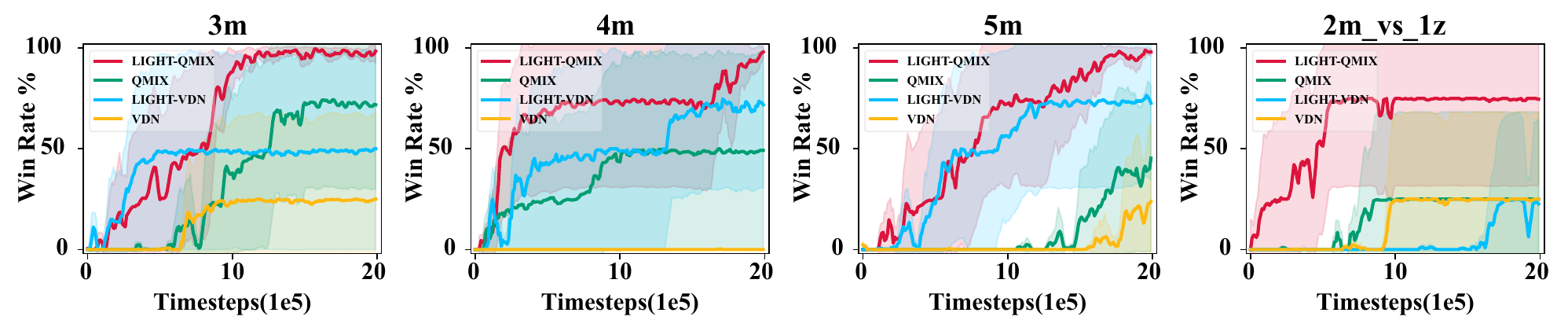}
    \vskip -0.1in
	\caption{LIGHT is plugged into two different baselines on sparse-reward scenarios.}
	\label{vdn_ablation}
    \vskip -0.2in
\end{figure*}

\begin{figure}[ht]
	\centering  
	\includegraphics[width=0.5\textwidth]{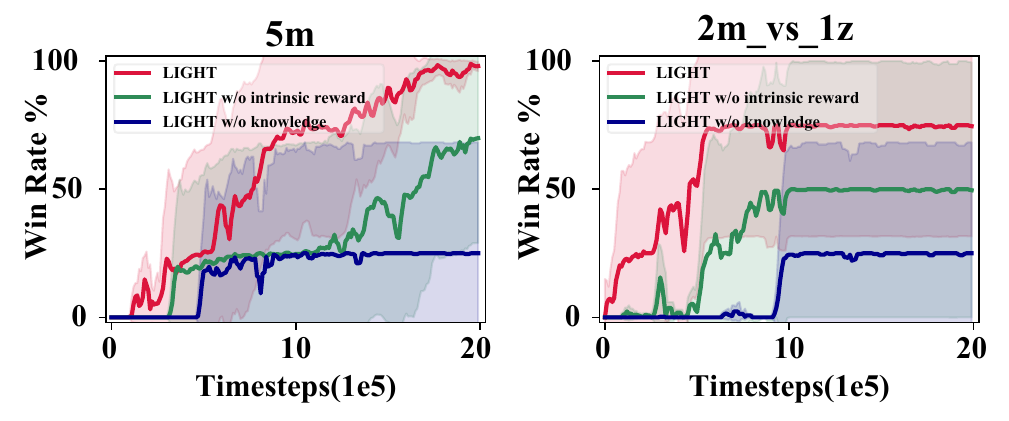}
    \vskip -0.15in
	\caption{The ablation study results of LIGHT, LIGHT w/o intrinsic reward, and LIGHT w/o knowledge on 5m and 2m\_vs\_1z scenarios.}
	\label{ablation}
    \vskip -0.1in
\end{figure}

\begin{figure}[ht]
	\centering
	\subfigure[Average step sizes]{
		\includegraphics[width=0.2\textwidth]{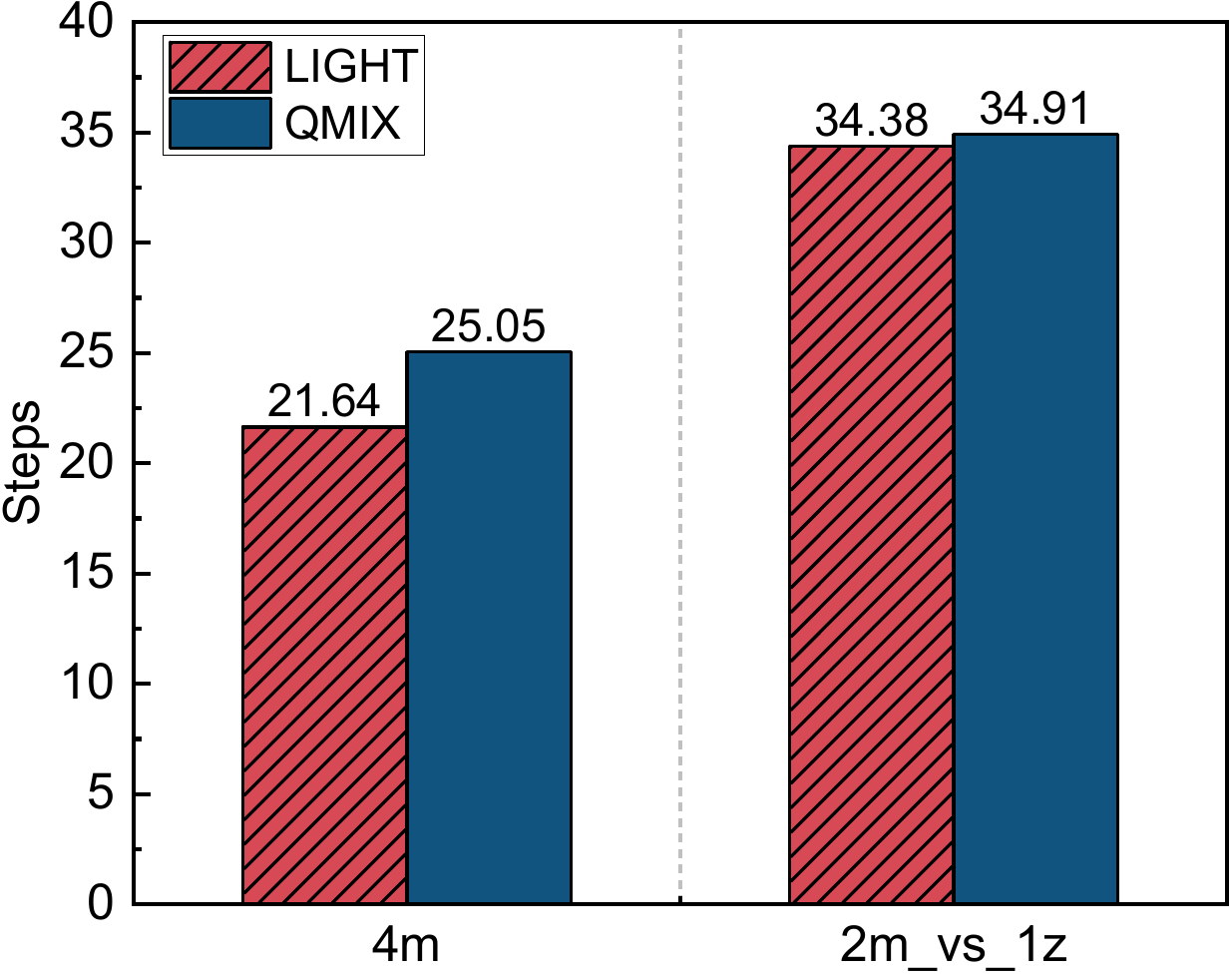}
	}
	\subfigure[Average similarity]{
		\includegraphics[width=0.2\textwidth]{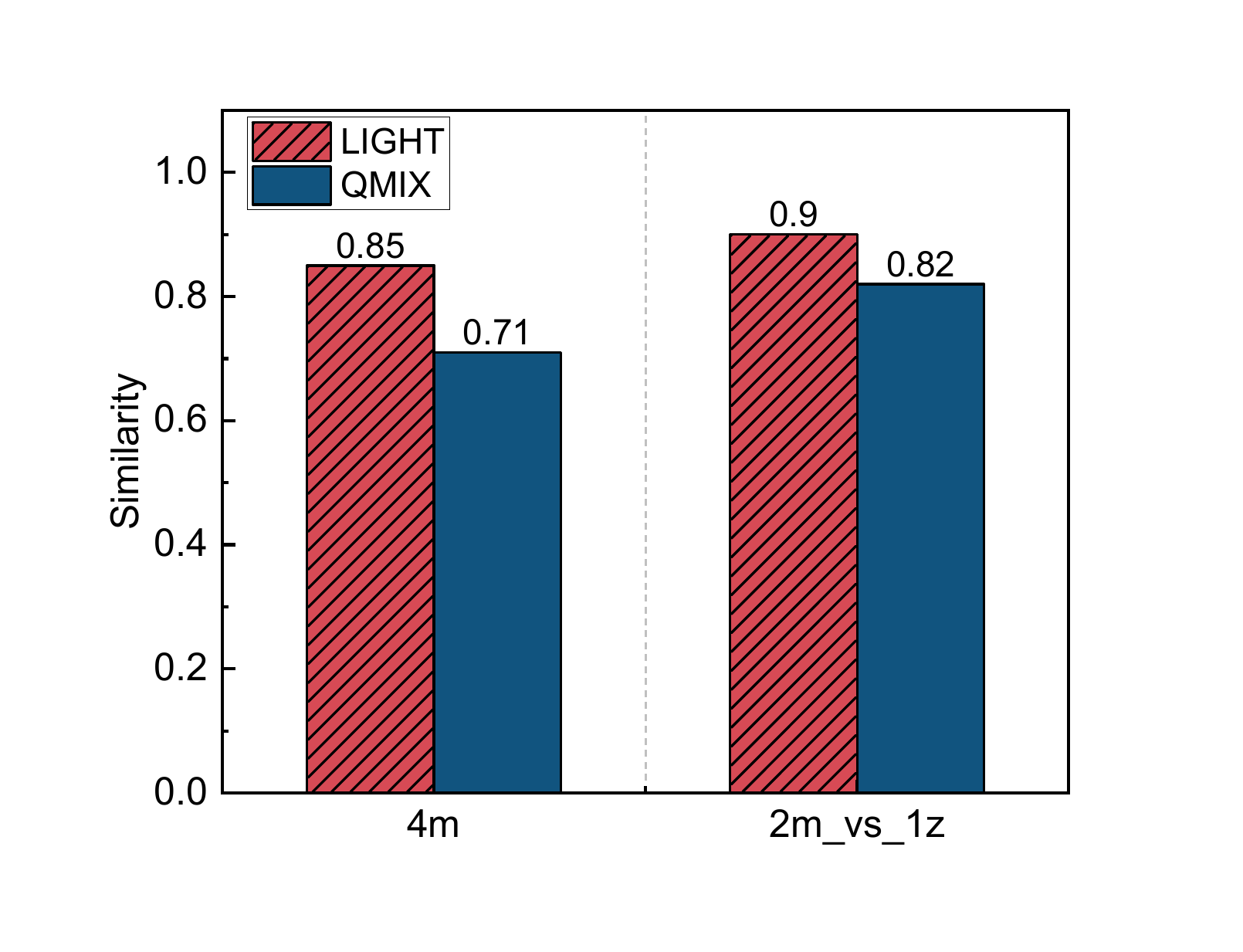}
	}
    \vskip -0.1in
	\caption{A comparison of alignment with human knowledge preference behaviors of LIGHT and QMIX on 4m and 2m\_vs\_1z maps.}
    \vskip -0.3in
	\label{similarity}
\end{figure}

\begin{figure}[ht]
	\centering
	\subfigure[{\fontsize{7}{1}\selectfont Intrinsic reward curves}]{
		\includegraphics[width=0.152\textwidth]{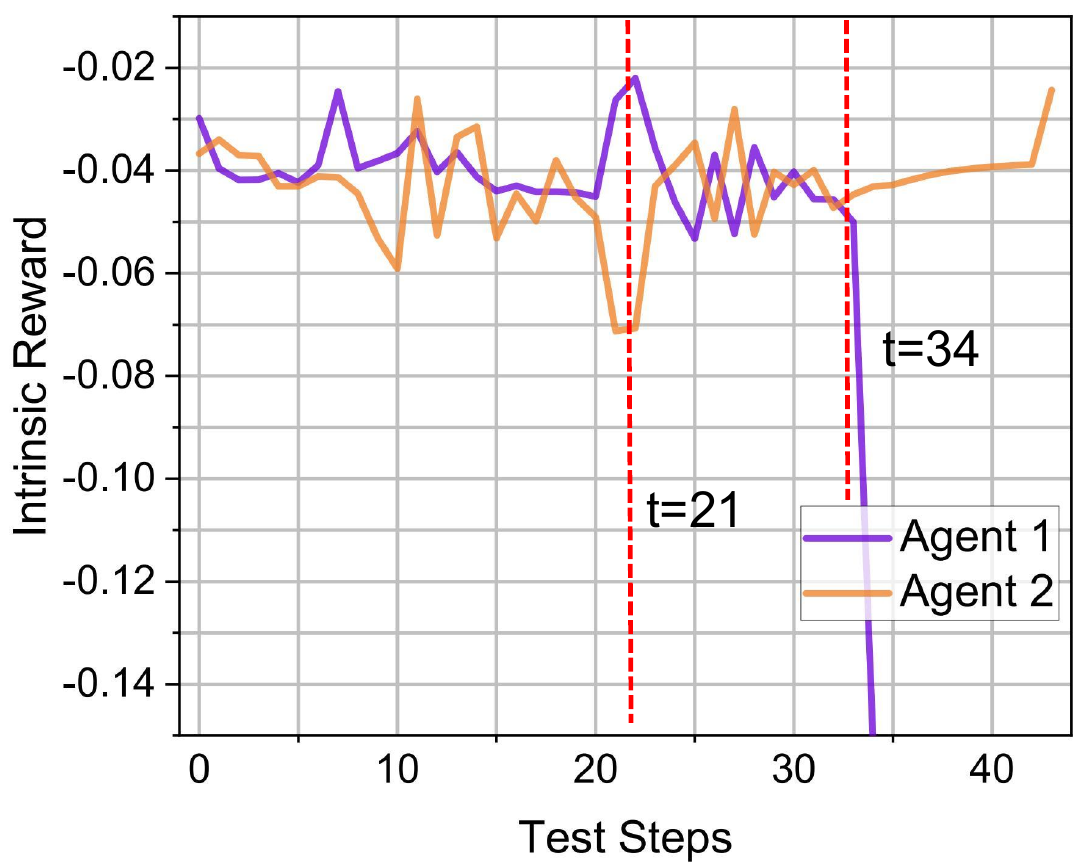}
	}
	\subfigure[t=21]{
		\includegraphics[width=0.14\textwidth]{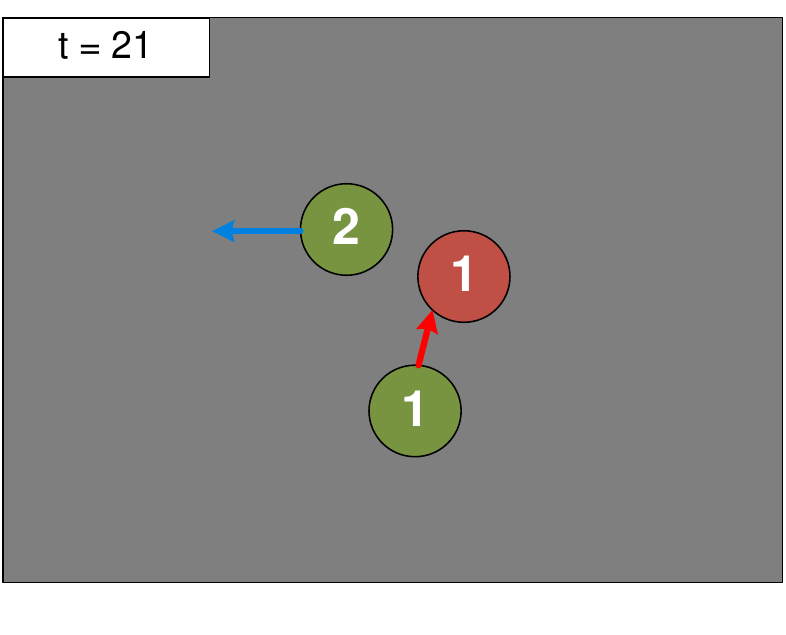}
	}
	\subfigure[t=34]{
		\includegraphics[width=0.14\textwidth]{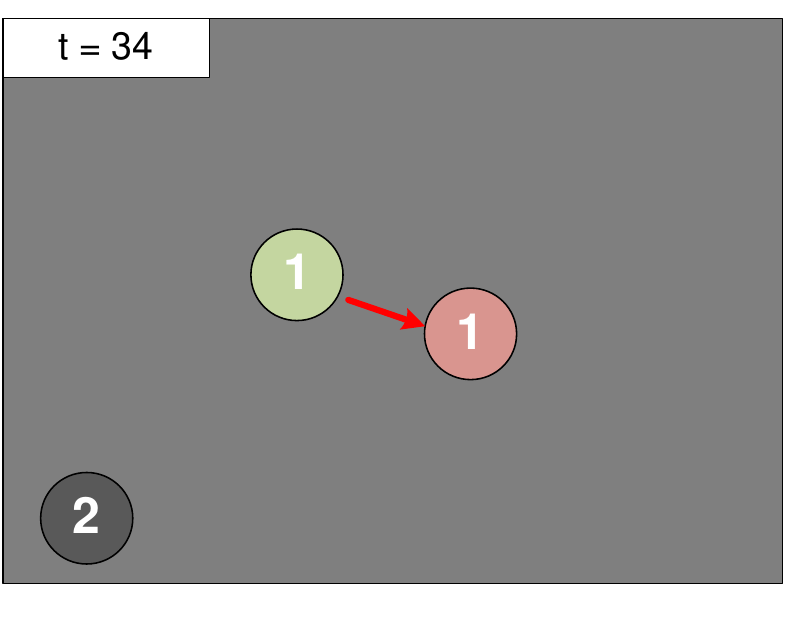}
	}
	\caption{An example of the intrinsic reward curves and auxiliary illustration on 2m\_vs\_1z map.
		Green circles and red circles represent allies and enemies, respectively, where the darker color indicates the higher health value of the agent.
		Gray circles indicate that the agent was killed.
		Blue arrows represent $move$, red arrows represent $attack$.}
	\label{2m_vs_1z_intrinsic_reward}
    \vspace{-0.5cm}
\end{figure}
\subsection{Overall Performance Comparison}
\textbf{Performance on LBF.} 
It shows in Fig.~\ref{performance_lbf} that LIGHT outperforms all baselines on constructed scenarios of LBF.
The poor performance of LIIR can be attributed to its failure to explore cooperative strategies.
Although MASER generates subgoals from the experience replay buffer, it fails to perform on all scenarios where rewards are highly sparse.
QMIX requires more steps to explore superior policies, which may be due to the lack of intrinsic rewards for each agent, resulting in an inability to guide agents toward efficient exploration with sparse rewards.
Despite alleviating QMIX's monotonic constraint, QTRAN still performs poorly due to its insufficient constraint relaxation in challenge scenarios. 
VDN yields competitive performance that spends fewer time steps to achieve a higher return than LIGHT before 0.5 steps.
However, it may be due to neglecting human expertise, which leads to an ultimately suboptimal performance compared with LIGHT.
In summary, LIGHT demonstrates impressive performance on effectiveness over all baselines, indicating that it is advantageous to consider human preferences when designing individual intrinsic rewards for MARL.

\textbf{Performance on SMAC.} 
We first ran all the considered algorithms on the conventional dense-reward setting.
As shown in Fig.~\ref{normal}, most algorithms can achieve near 100\% win rates on 3m, 4m, and 5m scenarios, and only LIIR can not work well on the 2m\_vs\_1z map.
It indicates that LIGHT and most algorithms are able to accomplish the tasks on scenarios with dense rewards.
Note that here we apply our LIGHT architecture to the individual Q-network of the QMIX and denote it as LIGHT.
We then evaluate all the algorithms on the difficult sparse-reward setting.
As shown in Fig.~\ref{sparse}, LIGHT performs better than all baselines on all scenarios.
QMIX achieves satisfactory performance, which may contribute to the efficient credit assignment for facilitating collaboration among agents.
QTRAN does not yield satisfactory performance, which may be due to the relaxation in practice that is insufficient for challenging domains.
Additionally, MASER performs better than VDN and LIIR, which should benefit from its individual intrinsic reward for each agent based on the actionable representation.
This can help agents reach their subgoals while maximizing the joint action value.
In summary, our method achieves impressive performance on all scenarios, demonstrating the advantage of LIGHT with attentive design through incorporating human expertise. 

We apply our LIGHT architecture to the individual Q-network of the ﬁne-tuned QMIX and VDN and denote them as LIGHT-QMIX and LIGHT-VDN, respectively.
As shown in Fig.~\ref{vdn_ablation}, our LIGHT-QMIX surpasses the ﬁne-tuned QMIX by a large margin in almost all scenarios, especially in 3m, 4m, 5m, and 2m\_vs\_1z scenarios.
Our LIGHT-VDN also signiﬁcantly improves the performance of the ﬁne-tuned VDN, and it even surpasses the ﬁne-tuned QMIX in most scenarios and achieves close performance to the LIGHT-QMIX, which minimizes the gaps between the VDN and QMIX algorithms.

\subsection{Visualizing the Learned Intrinsic Reward}
\textbf{Interpretability of LIGHT on LBF.}
We visualize these rewards to satisfy curiosity about how powerful the learned intrinsic reward function is for the policy learning.
As shown in Fig.~\ref{LBF}, we display some keyframes on 3-agent $\&$ 3-food map, where the red arrows represent the direction of movement.
As shown in Fig.~\ref{lbf-re}, we find that the intrinsic reward for agent-2 increases a lot, implying that this action is good at this time.
Fig.~\ref{lbf-t8} provides visualization evidence that agent-4 cannot eat food of a higher level than itself, which requires waiting to cooperate with other agents. 
At this stage, agent-2 moves north and cooperates with agent-4 to eat the food item in the next step.
As seen from Fig.~\ref{lbf-re}, at time step 10, we see that
agent-1 and agent-3 achieve the supreme intrinsic rewards compared to others, which indicates that they all select great behaviors in this step.
In Fig.~\ref{lbf-t10}, agent-3 selects the $East$ action, and then agent-1 and agent-3 form a cooperative alliance to jointly acquire the food item exceeding their individual capability levels.
The above analysis shows the superiority of the designed intrinsic reward function that plays a pivotal role in providing important feedback for agents, enabling real-time behavior evaluation and policy optimization in complex cooperative or competitive environments. 

\textbf{Interpretability of LIGHT on SMAC.}
To better understand the impact of the learned intrinsic reward function on policy training, we propose a direct visualization of these rewards. 
Specifically, we plot the intrinsic rewards assigned to each agent at every step of a complete trajectory during testing.
It is important to note that the intrinsic rewards do not influence the learned policy and are not utilized in trajectory generation.
For better visualization and clarity, we select two test replays from scenarios 4m and 2m\_vs\_1z and chart the intrinsic rewards for all agents involved.
Figures 4 and \ref{2m_vs_1z_intrinsic_reward} illustrate these intrinsic rewards in the 4m and 2m\_vs\_1z scenarios, respectively.

In Fig.~\ref{4m_intrinsic_reward}-(a), at time step 9, the intrinsic reward of agent-1 rises to near $0$ because it has the lowest health and chooses $move$ rather than $attack$ as revealed in Fig.~\ref{4m_intrinsic_reward}-(b). 
It indicates that the $attack$ action is not a good behavior for agent-1 at this time.
After time step 18, agent-2 has been attacked until it dies, and it receives a large negative intrinsic reward.

On the 2m\_vs\_1z map, we find that agent-1 receives a larger reward at time step 21 in Fig.~\ref{2m_vs_1z_intrinsic_reward}-(a). 
This may be because it takes on more of the task of drawing fire and attacking at moments of higher health than its companions, while agent-2 stops firing and falls behind its team with a lower intrinsic reward.
After several steps, agent-2 obtains cooperative skills, and the intrinsic rewards gradually increase. 
Meanwhile, agent-1 is killed and receives a lower intrinsic reward.
By visualizing the intrinsic reward curves for the two maps, the results illustrate that the generated intrinsic reward can effectively provide diverse feedback signals, which are highly informative in assessing the agents' immediate behaviors.

\subsection{Ablation Studies}
To understand the impact of each component of LIGHT, we conduct ablation studies to answer the following questions:
(1) How does the intrinsic reward influence performance?
(2) How does human knowledge influence performance?
To study components (1) and (2), LIGHT w/o intrinsic reward represents removing intrinsic rewards, and LIGHT w/o human knowledge represents human knowledge with randomly generated distributions, respectively.
We carry out ablation studies on 5m and 2m\_vs\_1z maps.
As shown in Fig.~\ref{ablation}, the ablation of each part of LIGHT brings a noticeable decrease in performance.
Specifically, the performance of LIGHT w/o human knowledge decreases, which indicates that human knowledge is beneficial in guiding the agent to explore better.
Besides, the performance of LIGHT w/o intrinsic reward is lower than LIGHT, which indicates that the intrinsic rewards can ultimately induce better exploration in sparse-reward environments.
To summarize, LIGHT, conditioned on all components, gives the best performance, which could improve exploration with the given limited learning time steps.

\subsection{Behavior analysis}
In addition to evaluating the performance of LIGHT, we are more curious about whether the behavior aligns with the given human knowledge.
To study how human knowledge influences the behavior of LIGHT, we compare the consistent behavior of LIGHT and QMIX on 4m and 2m\_vs\_1z maps with sparse-reward settings.
Specifically, we make a statistical analysis of whether the action of the agent is consistent with the given human knowledge at each time step during testing for $100$ episodes.
Here, we consider the behavior to be consistent if the agent produces an action that is consistent with human knowledge.
On the 4m and 2m\_vs\_1z maps, as shown in Fig.~\ref{similarity}-(a), we can find that the average steps of LIGHT are $21.64$ and $34.38$ over $100$ episodes, while QMIX requires $25.05$ and $34.91$ time steps, respectively.
It indicates that LIGHT requires fewer time steps to solve the task with a more optimal policy.
As shown in Fig.~\ref{similarity}-(b), LIGHT has more behaviors aligning with the given human knowledge than QMIX across both scenarios, which indicates that our method can efficiently capture the given human knowledge to facilitate the learning process.
The above case studies demonstrate that LIGHT can not only promote learning efficiency but also provide a novel method to incorporate the given human preference into the behavior of the agents.

\section{Conclusion}
In this work, we propose a novel value decomposition framework called LIGHT to leverage human expertise to accelerate the learning process of MARL agents.
LIGHT produces intrinsic rewards to induce the agent to explore efficiently by considering both each agent's action distribution and human preference at an early stage.
This end-to-end framework can be combined with existing value decomposition algorithms to deal with the sparse-reward setting tasks.
Experiments on the challenging LBF and SMAC benchmarks show that our method obtains the best performance on almost all sparse-reward maps, and the intrinsic reward module of LIGHT can help the behavior of agents better align with human preferences.
In the future, this simple yet effective method further motivates us to explore more effective ways to utilize intrinsic rewards by incorporating human knowledge in more challenging tasks.

%\section*{Acknowledgements}
%This work was supported in part by the National Key Research and Development Program of China under Grant 2023YFD2001003 and the Fundamental Research Funds for the Central Universities under Grant 011814380048.

\bibliographystyle{IEEEtran}
\bibliography{mybibfile}

\begin{thebibliography}{10}
\providecommand{\url}[1]{#1}
\csname url@rmstyle\endcsname
\providecommand{\newblock}{\relax}
\providecommand{\bibinfo}[2]{#2}
\providecommand\BIBentrySTDinterwordspacing{\spaceskip=0pt\relax}
\providecommand\BIBentryALTinterwordstretchfactor{4}
\providecommand\BIBentryALTinterwordspacing{\spaceskip=\fontdimen2\font plus
\BIBentryALTinterwordstretchfactor\fontdimen3\font minus \fontdimen4\font\relax}
\providecommand\BIBforeignlanguage[2]{{%
\expandafter\ifx\csname l@#1\endcsname\relax
\typeout{** WARNING: IEEEtran.bst: No hyphenation pattern has been}%
\typeout{** loaded for the language `#1'. Using the pattern for}%
\typeout{** the default language instead.}%
\else
\language=\csname l@#1\endcsname
\fi
#2}}

\bibitem{car}
Y.~Cao, W.~Yu, W.~Ren, and G.~Chen, ``An overview of recent progress in the study of distributed multi-agent coordination,'' \emph{IEEE Trans. Industr. Inform.}, pp. 427--438, 2012.

\bibitem{zhang2011coordinated}
C.~Zhang and V.~Lesser, ``Coordinated multi-agent reinforcement learning in networked distributed pomdps,'' in \emph{Proceedings of the AAAI Conference on Artificial Intelligence}, 2011.

\bibitem{ghorbani2019interpretation}
A.~Ghorbani, A.~Abid, and J.~Zou, ``Interpretation of neural networks is fragile,'' in \emph{Proceedings of the AAAI Conference on Artificial Intelligence}, 2019, pp. 3681--3688.

\bibitem{huttenrauch2017guided}
M.~H{\"u}ttenrauch, A.~{\v{S}}o{\v{s}}i{\'c}, and G.~Neumann, ``Guided deep reinforcement learning for swarm systems,'' \emph{arXiv:1709.06011}, 2017.

\bibitem{qpd}
Y.~Yang, J.~Hao, G.~Chen, H.~Tang, Y.~Chen, Y.~Hu, C.~Fan, and Z.~Wei, ``Q-value path decomposition for deep multiagent reinforcement learning,'' in \emph{Proceedings of the International Conference on Machine Learning}, 2020, pp. 10\,706--10\,715.

\bibitem{rode}
T.~Wang, T.~Gupta, A.~Mahajan, B.~Peng, S.~Whiteson, and C.~Zhang, ``{RODE}: Learning roles to decompose multi-agent tasks,'' in \emph{Proceedings of the International Conference on Learning Representations}, 2020, pp. 1--20.

\bibitem{roma}
T.~Wang, H.~Dong, V.~Lesser, and C.~Zhang, ``{ROMA}: Multi-agent reinforcement learning with emergent roles,'' in \emph{Proceedings of the International Conference on Machine Learning}, 2020, pp. 9876--9886.

\bibitem{vdn}
P.~Sunehag, G.~Lever, A.~Gruslys, W.~M. Czarnecki, V.~Zambaldi, M.~Jaderberg, M.~Lanctot, N.~Sonnerat, J.~Z. Leibo, K.~Tuyls, \emph{et~al.}, ``Value-decomposition networks for cooperative multi-agent learning based on team reward,'' in \emph{Proceedings of the International Conference on Autonomous Agents and MultiAgent Systems}, 2018, pp. 2085--2087.

\bibitem{qmix}
T.~Rashid, M.~Samvelyan, C.~Schroeder, G.~Farquhar, J.~Foerster, and S.~Whiteson, ``{QMIX}: Monotonic value function factorisation for deep multi-agent reinforcement learning,'' in \emph{Proceedings of the International Conference on Machine Learning}, 2018, pp. 4295--4304.

\bibitem{qplex}
J.~Wang, Z.~Ren, T.~Liu, Y.~Yu, and C.~Zhang, ``{QPLEX}: Duplex dueling multi-agent {Q}-learning,'' in \emph{Proceedings of the International Conference on Learning Representations}, 2020, pp. 1--27.

\bibitem{mixrts}
Z.~Liu, Y.~Zhu, Z.~Wang, Y.~Gao, and C.~Chen, ``Mixrts: Toward interpretable multi-agent reinforcement learning via mixing recurrent soft decision trees,'' \emph{IEEE Transactions on Pattern Analysis and Machine Intelligence}, vol.~47, no.~5, pp. 4090--4107, 2025.

\bibitem{liu23be}
Z.~Liu, Y.~Zhu, and C.~Chen, ``{N}$\text{{A}}^\text{2}${Q}: Neural attention additive model for interpretable multi-agent q-learning,'' in \emph{Proceedings of the International Conference on Machine Learning}, vol. 202, 2023, pp. 22\,539--22\,558.

\bibitem{busoniu2008comprehensive}
L.~Busoniu, R.~Babuska, and B.~De~Schutter, ``A comprehensive survey of multiagent reinforcement learning,'' \emph{IEEE Transactions on Systems, Man, and Cybernetics}, vol.~38, no.~2, pp. 156--172, 2008.

\bibitem{sadeghlou2014dynamic}
M.~Sadeghlou, M.~R. Akbarzadeh-T, and M.~B. Naghibi-S, ``Dynamic agent-based reward shaping for multi-agent systems,'' in \emph{Iranian Conference on Intelligent Systems}, 2014, pp. 1--6.

\bibitem{wong2021multiagent}
A.~Wong, T.~B{\"a}ck, A.~V. Kononova, and A.~Plaat, ``Multiagent deep reinforcement learning: Challenges and directions towards human-like approaches,'' \emph{arXiv:2106.15691}, 2021.

\bibitem{liu2021cooperative}
I.-J. Liu, U.~Jain, R.~A. Yeh, and A.~Schwing, ``Cooperative exploration for multi-agent deep reinforcement learning,'' in \emph{Proceedings of the International conference on machine learning}, 2021, pp. 6826--6836.

\bibitem{mahajan2019maven}
A.~Mahajan, T.~Rashid, M.~Samvelyan, and S.~Whiteson, ``Maven: Multi-agent variational exploration,'' \emph{Advances in neural information processing systems}, vol.~32, 2019.

\bibitem{liir}
Y.~Du, L.~Han, M.~Fang, J.~Liu, T.~Dai, and D.~Tao, ``{Liir}: Learning individual intrinsic reward in multi-agent reinforcement learning,'' in \emph{Advances in neural information processing systems}, vol.~32, 2019.

\bibitem{maser}
J.~Jeon, W.~Kim, W.~Jung, and Y.~Sung, ``{MASER}: Multi-agent reinforcement learning with subgoals generated from experience replay buffer,'' in \emph{Proceedings of the International Conference on Machine Learning}, 2022, pp. 10\,041--10\,052.

\bibitem{wang2022individual}
L.~Wang, Y.~Zhang, Y.~Hu, W.~Wang, C.~Zhang, Y.~Gao, J.~Hao, T.~Lv, and C.~Fan, ``Individual reward assisted multi-agent reinforcement learning,'' in \emph{Proceedings of the International Conference on Machine Learning}.\hskip 1em plus 0.5em minus 0.4em\relax PMLR, 2022, pp. 23\,417--23\,432.

\bibitem{xu2023haven}
Z.~Xu, Y.~Bai, B.~Zhang, D.~Li, and G.~Fan, ``Haven: hierarchical cooperative multi-agent reinforcement learning with dual coordination mechanism,'' in \emph{Proceedings of the AAAI Conference on Artificial Intelligence}, vol.~37, no.~10, 2023, pp. 11\,735--11\,743.

\bibitem{xu2020hierarchical}
X.~Xu, T.~Huang, P.~Wei, A.~Narayan, and T.-Y. Leong, ``Hierarchical reinforcement learning in starcraft ii with human expertise in subgoals selection,'' \emph{arXiv preprint arXiv:2008.03444}, 2020.

\bibitem{fischer2019dl2}
M.~Fischer, M.~Balunovic, D.~Drachsler-Cohen, T.~Gehr, C.~Zhang, and M.~Vechev, ``Dl2: training and querying neural networks with logic,'' in \emph{Proceedings of the International Conference on Machine Learning}, 2019, pp. 1931--1941.

\bibitem{9403986}
Y.~Zhu, Z.~Wang, C.~Chen, and D.~Dong, ``Rule-based reinforcement learning for efficient robot navigation with space reduction,'' \emph{IEEE/ASME Transactions on Mechatronics}, vol.~27, no.~2, pp. 846--857, 2022.

\bibitem{9970401}
Y.~Zhu, X.~Yin, and C.~Chen, ``Extracting decision tree from trained deep reinforcement learning in traffic signal control,'' \emph{IEEE Transactions on Computational Social Systems}, vol.~10, no.~4, pp. 1997--2007, 2023.

\bibitem{zhang2020kogun}
P.~Zhang, J.~Hao, W.~Wang, H.~Tang, Y.~Ma, Y.~Duan, and Y.~Zheng, ``Kogun: accelerating deep reinforcement learning via integrating human suboptimal knowledge,'' \emph{arXiv preprint arXiv:2002.07418}, 2020.

\bibitem{qtran}
K.~Son, D.~Kim, W.~J. Kang, D.~E. Hostallero, and Y.~Yi, ``{QTRAN}: Learning to factorize with transformation for cooperative multi-agent reinforcement learning,'' in \emph{Proceedings of the International Conference on Machine Learning}, 2019, pp. 5887--5896.

\bibitem{ICQL}
S.~W. Wendelin~Böhmer, Tabish~Rashid, ``Exploration with unreliable intrinsic reward in multi-agent reinforcement learning,'' in \emph{Proceedings of the International Conference on Machine Learning}, 2019.

\bibitem{offline}
F.~Zhang, C.~Jia, Y.-C. Li, L.~Yuan, Y.~Yu, and Z.~Zhang, ``Discovering generalizable multi-agent coordination skills from multi-task offline data,'' in \emph{Proceedings of the International Conference on Learning Representations}, 2023.

\bibitem{lbf}
F.~Christianos, L.~Sch{\"a}fer, and S.~Albrecht, ``Shared experience actor-critic for multi-agent reinforcement learning,'' in \emph{Advances in Neural Information Processing Systems}, 2020, pp. 10\,707--10\,717.

\bibitem{smac}
M.~Samvelyan, T.~Rashid, C.~Schroeder~de Witt, G.~Farquhar, N.~Nardelli, T.~G. Rudner, C.-M. Hung, P.~H. Torr, J.~Foerster, and S.~Whiteson, ``{The StarCraft Multi-Agent Challenge},'' in \emph{Proceedings of the International Conference on Autonomous Agents and MultiAgent Systems}, 2019, pp. 2186--2188.

\end{thebibliography}

\end{document}